\documentclass{article}
\usepackage{arxiv}

\usepackage[T1]{fontenc}
\usepackage[utf8]{inputenc}
\usepackage{amsmath,amssymb}
\usepackage{graphicx}
\usepackage{booktabs}
\usepackage{tabularx}
\usepackage{array}
\usepackage{multirow}
\usepackage{enumitem}
\usepackage[numbers,sort&compress]{natbib}
\usepackage[hidelinks]{hyperref}

\renewcommand{\shorttitle}{SAILS for PSMA PET/CT Segmentation}
\title{\textbf{Spatial Atlas-Informed Loss Ensembles for Whole-Body PSMA PET/CT Metastasis Segmentation}}
\author{%
  \begin{minipage}{0.98\textwidth}
  \centering\normalfont
  \textbf{Grant N. Chau}$^{1,2,*}$ \quad
  \textbf{Bashirul A. Biswas}$^{1}$ \quad
  \textbf{Biratal Wagle}$^{1}$ \\[4pt]
  \textbf{Matthew E. Maeder}$^{3}$ \quad
  \textbf{Michael S. Leapman}$^{4}$ \quad
  \textbf{James B. Yu}$^{5}$ \\[4pt]
  \textbf{Indrani Bhattacharya}$^{1,2}$ \\[8pt]
  {\small $^{1}$Department of Computer Science, Dartmouth College, Hanover, NH, USA}\\
  {\small $^{2}$Department of Biomedical Data Science, Geisel School of Medicine at Dartmouth, Hanover, NH, USA}\\
  {\small $^{3}$Radiology, Dartmouth Hitchcock Medical Center, Lebanon, NH, USA}\\
  {\small $^{4}$Department of Urology, Yale School of Medicine, New Haven, CT, USA}\\
  {\small $^{5}$Radiation Oncology, Dartmouth Hitchcock Medical Center, Lebanon, NH, USA}\\[5pt]
  {\small $^{*}$Corresponding author: \texttt{grant.n.chau.gr@dartmouth.edu}; ORCID: 0009-0000-5836-3902}
  \end{minipage}%
}
\date{}

\begin{document}
\maketitle

\begin{abstract}
\textbf{Background:} Existing whole-body PSMA PET/CT prostate metastasis segmentation methods generally rely on images alone, without explicitly incorporating where metastases tend to occur. We investigate whether a population metastasis atlas can improve lesion detection when integrated into model training. \textbf{Methods:} We constructed voxel-wise mean and standard deviation maps from 375 expert-annotated training scans and used them to weight the cross-entropy loss of an nnU-Net. The additional penalties for missed lesions and false positives depend on spatial atlas statistics and diminish as predictions become confidently correct. Seven configurations were compared on validation data; three were combined by majority voting into SAILS (\textbf{S}patial \textbf{A}tlas-\textbf{I}nformed \textbf{L}oss en\textbf{S}emble). We compared SAILS with a baseline nnU-Net and inference-time atlas fusion on AutoPET~III (110 disease-positive cases) and Deep-PSMA (100 external cases). No atlas or registration is needed at inference. \textbf{Results:} On AutoPET~III, SAILS improved whole-body Dice (0.531 vs.\ 0.520; $p=0.006$), reduced missed-lesion volume (9.5 vs.\ 12.1~cc/case; $p<0.01$), and increased pooled lesion sensitivity (0.857 vs.\ 0.843; $p=0.004$). On Deep-PSMA, SAILS detected 126 additional reference lesions (sensitivity 0.769 vs.\ 0.751; $p<0.001$), without a significant whole-body Dice change ($-0.009$; $p=0.36$). High-risk bone Dice improved in both cohorts (0.576 vs.\ 0.552 and 0.667 vs.\ 0.653; both $p<0.01$). False-positive volume increased by 2.3 and 18.2~cc/case, respectively ($p<0.001$). \textbf{Conclusions:} Atlas-weighted training improves lesion detection, especially in high-risk bone, without test-time atlas registration, but increases false positives.
\end{abstract}


\paragraph{Highlights.}
\begin{itemize}[leftmargin=1.5em,topsep=2pt,itemsep=1pt]
\item A data-driven whole-body prostate metastasis atlas integrated into segmentation training via a novel atlas-weighted loss function.
\item A training-free majority vote of atlas-weighted networks (SAILS) increases lesion-level sensitivity.
\item Improved segmentation of high-risk bone metastases on an internal test set and an external cohort.
\item No atlas or registration is needed at inference.
\end{itemize}

\section{Introduction}\label{sec:intro}

Prostate cancer is the second leading cause of cancer death among men in the United States, and this mortality is driven almost entirely by metastatic disease. Five-year relative survival falls from nearly 100\% for localized disease to roughly 30\% once distant metastases are present \citep{ACSsurvival}. The number of these men living with metastatic prostate cancer is projected to keep growing \citep{Desai22,Devasia23}. Bone is involved in roughly 84\% of men with metastatic prostate cancer. The liver is involved in only about 10\%, but liver involvement carries the worst prognosis \citep{Gandaglia14,Kadeerhan23,Shiner24}.

Clinical management and treatment of metastatic prostate cancer is personalized and depends on where and how extensively the disease has spread. For oligometastatic disease, generally defined as up to five metastatic sites, radiotherapy directed at every identified metastasis improves progression-free survival \citep{phillips2020outcomes,alberto2022role}. For bone metastases at high risk of skeletal-related events (for example, bulky lesions or lesions at weight-bearing joints or the junctional spine), prophylactic radiotherapy reduced skeletal-related events and hospitalization and prolonged overall survival in a randomized phase II trial \citep{Gillespie24HighRiskBone}. For more widespread PSMA-positive disease, the radioligand therapy $^{177}$Lu-PSMA-617 improved overall survival in the VISION trial and PSA response compared with cabazitaxel in the TheraP trial. Notably, both trials used PSMA PET to select patients \citep{Sartor21VISION,Hofman21TheraPResults}. $^{177}$Lu-PSMA-617 was first approved for metastatic castration-resistant prostate cancer, and in 2026 its approval was extended, in combination with androgen receptor pathway inhibitor therapy, to PSMA-positive metastatic hormone-sensitive prostate cancer \citep{FDA26Pluvicto}. Each of these treatment decisions requires knowing the number, location, and extent of metastases. PSMA PET/CT is more accurate than conventional imaging for this purpose and therefore is increasingly used for staging, treatment selection, and response assessment\citep{Hofman20proPSMA}.

Lesion segmentation on PSMA PET/CT also enables more fine-grained quantitative measures of disease burden within a patient. Whole-body and organ-specific PSMA-positive tumor volumes are prognostic for overall survival \citep{Seifert20TTV,Seifert23RiskScore,Zang25AIVolume} and are inputs to nomograms and multivariable models that predict survival outcomes after $^{177}$Lu-PSMA-617 therapy \citep{Gafita21Nomograms,Herrmann24Multivariable}. Of course, these metrics are only as reliable as the underlying segmentation process can guarantee \citep{Kim23Thresholding}.

For clinicians, the problem of identifying every metastatic site on whole-body PSMA PET/CT remains challenging. Metastases are frequently small and of low uptake, and PSMA uptake is not even specific to cancer. Sympathetic ganglia and inflammatory lymph nodes can mimic nodal metastases, benign bone findings such as fractures, hemangiomas, and Paget's disease can mimic bone metastases, and nonspecific bone uptake is common with \textsuperscript{18}F-labeled PSMA tracers, particularly \textsuperscript{18}F-PSMA-1007,\citep{Jochumsen24}. The high physiological liver uptake of \textsuperscript{18}F-PSMA-1007 may also reduce sensitivity for liver metastases \citep{Jochumsen24}, which carry the worst prognosis. Inter-reader agreement is only moderate ($\kappa=0.41$--$0.76$), and both missed lesions and false positives are common \citep{Donswijk24,Miksch20}. A patient may have dozens of metastatic foci, which places a large cognitive burden on even expert clinicians and readers. Meanwhile, the volume of PSMA PET/CT is growing, and the increase in trained nuclear medicine specialists does not match this ever-growing rate \citep{AfshariMirak25,Medicus25}. Thus, reliable and automated lesion segmentation could reduce reading burden and variability and could make the quantitative burden measures available in routine practice.

Automated whole-body lesion segmentation has advanced substantially in recent years. The self-configuring nnU-Net framework \citep{Isensee21} has become a strong and widely used baseline, and the public AutoPET-III challenge, which includes manually annotated PSMA PET/CT has created multi-center benchmarks for this task \citep{Gatidis24,Jeblick26}. The top performing AutoPET III methods are built on nnU-Net with larger residual encoders, pretraining, and auxiliary organ segmentation \citep{Rokuss24,Kalisch24}. Dedicated PSMA PET/CT tools include early 2.5D (analyzing adjacent 2-dimensional slices) U-Net models \citep{Zhao20}, nnU-Net-based systems trained on multicenter cohorts \citep{Assadi24}, and the aPROMISE software \citep{Nickols22,Johnsson22}. aPROMISE segments organs on low-dose CT with deep learning and then detects candidate lesions on PET with site-specific blob detectors, after suppressing modeled normal uptake in the liver, kidneys and bladder. Candidates are then filtered by SUV$_{\max}$, SUV$_{\mathrm{mean}}$, volume and location, and segmented with the fast marching method \citep{Sethian96FastMarching}, which grows a region outward from each detected blob \citep{Johnsson22}. Lesion detection therefore relies on uptake-based rules rather than a learned lesion model. In a reader study, it supported consistent and standardized reporting \citep{Nickols22}. These methods segment high-PET-SUV (Standardized Uptake Value) and larger lesions well. Despite this progress, these models remain error-prone in several clinically relevant ways. Critically, they miss lesions at rare but high-risk sites such as the liver, which is a region where lesions are underrepresented in training data and physiological uptake reduces contrast \citep{RiosSanchez25PSA50,Jochumsen24}. They also often miss bone lesions, small and low-SUV lesions. They produce false positives at sites of intense physiological uptake where metastases are implausible.

All of these methods use image intensities and anatomic reference from CT scans alone for segmentation, without considering clinical domain knowledge about prostate cancer spread. Clinicians, however, read PSMA PET/CT with their prior domain knowledge of how prostate cancer spreads. They expect disease in the pelvic nodes and axial skeleton, discount intense uptake in organs where metastases are rare, and pay especially close attention to sites where a missed lesion would change clinical outcomes. Prior work has used anatomical context such as organ labels as an auxiliary task \citep{Kalisch24} or CT organ segmentation to contextualize tracer uptake \citep{Johnsson22}. However, to the best of our knowledge, no previous whole-body PSMA PET/CT segmentation method has used a population-level metastasis atlas to guide the training loss.

Spatial priors have improved detection and segmentation in other settings. In brain MRI, probabilistic atlases of tissue classes have been combined with image intensities within a single generative model \citep{Ashburner05}. Learned anatomical priors have also allowed convolutional networks to segment brain MRI without paired image--label training data \citep{Dalca18}. In prostate MRI, population cancer atlases have been used to guide diffusion-based cancer detection \citep{LiProstAtlasDiff25}, and zonal priors have improved the detection of aggressive prostate cancer \citep{Bhattacharya22zonal}. For metastatic spread, PET-based lymph node atlases have mapped where positive nodes occur in order to guide radiotherapy target volumes \cite{Hegemann16NodalAtlas,Schiller21NodalAtlas}. These atlases cover only the pelvic and para-aortic nodes, and they have not been used in segmentation models. A whole-body atlas of prostate cancer metastases is harder to build, as it needs voxel-wise lesion annotation for many whole-body scans and deformable registration across patients who differ widely in body physique, posture, and field of view, and metastases occupy only a tiny fraction of the body.

In our previous work, MAPLE (Metastasis Atlas for Pooled Log-odds Estimation) \citep{ChauAtlasFusion26}, we built a whole-body metastasis atlas from 375 expert-annotated PSMA PET/CT scans and fused it with a trained nnU-Net's output at inference time. This improved Dice on an in-distribution test set and on an independent cohort with no retraining. However, inference-time fusion has two inherent limitations. First, it can only rescale predictions that the network has already made and cannot change what the network has learned. A lesion assigned near-zero probability, or a physiologically common spot confidently called a lesion is therefore difficult for the prior to correct. Second, the atlas must be deformably registered to every test patient. In our pipeline this takes about 20 minutes per case, compared to 1.25 minutes for nnU-Net inference. Registration also requires a CT of adequate quality and field of view and can fail on truncated scans or atypical anatomy.

An alternative is to inject the prior through the training loss, which changes what the network learns without changing its architecture or inference cost. Custom loss functions are a well-established paradigm for guiding segmentation networks \citep{Ma21LossOdyssey}. Dice-based and class-weighted losses counter the imbalance between small lesions and the much larger background voxels \citep{Milletari16}. The focal loss down-weights voxels the network already classifies correctly, so that training focuses on hard voxels \citep{Lin17Focal}. Tversky-type losses penalize false negatives and false positives by different amounts \citep{Salehi17Tversky,Abraham19FocalTversky}. In PSMA PET/CT specifically, a size-weighted batch-wise Dice loss improved the detection of small lesions \citep{Xu23}. These weightings depend on lesion or class properties (e.g. size, intensity, class frequency) or vary spatially only through image-derived maps such as the boundary weight map of the original U-Net \citep{Ronneberger15}. A voxel in the skeleton is therefore treated identically to a voxel in the liver despite their very different metastatic prevalence and prognostic weight. Atlas-derived loss weights have been explored for organ segmentation, where a liver probability atlas reweights the cross-entropy toward voxels with an ambiguous prior \citep{Zheng19}. To the best of our knowledge, however, a population prior of where metastases occur has not been used to modulate the training loss of a whole-body PET/CT lesion segmentation network.

We therefore propose to bring a whole-body metastasis atlas into training through a novel atlas-weighted loss function (Fig.~\ref{fig:overview}). The proposed atlas-weighted loss penalizes errors more heavily in regions where metastases are likely to occur and regions that demonstrate variability in metastases distribution, and this extra penalty fades as the network becomes confident and correct. The atlas affects training only, so the resulting network is architecturally identical to a standard nnU-Net and needs no atlas or registration at inference. Our contributions are:

\begin{itemize}

    \item We extend the whole-body, population-level spatial atlas of PSMA-avid metastases from our prior work \citep{ChauAtlasFusion26}, built following the population-atlas strategy of whole-body MRI \citep{Starck24}, with a voxel-wise standard deviation map that captures where metastatic occurrence is variable. Spatial priors in prostate cancer imaging have so far been organ-specific \citep{LiProstAtlasDiff25,Bhattacharya22zonal} or limited to lymph nodes \citep{Hegemann16NodalAtlas,Schiller21NodalAtlas}.
    \item We develop an atlas-weighted loss that penalizes missed lesions and false positives more heavily where metastases occur across patients, which includes common sites such as bone and rare high-risk sites such as the liver. The loss reduces to standard cross-entropy where the atlas carries no signal or the network is already confident and correct.
    \item We combine three atlas-weighted networks by majority voting into \textbf{SAILS} (\textbf{S}patial \textbf{A}tlas-\textbf{I}nformed \textbf{L}oss en\textbf{S}emble). The members and the voting rule are selected on the validation set by requiring higher Dice together with fewer missed and false-positive voxels. SAILS needs no retraining and no atlas or registration at inference.
    \item We perform a comprehensive, clinically relevant evaluation at the voxel, lesion, and anatomic-region level, including metastases in bone, high-risk bone, and liver. We find that SAILS improves detection (missed-lesion volume, lesion sensitivity, and segmentation of high-risk bone metastases), while increasing false-positive volume.
    \item We evaluate generalizability on a held-out AutoPET~III test set and the external Deep-PSMA cohort \citep{DeepPSMA25}, which differ by scanner, tracer, resolution, annotation protocol, and disease burden.
\end{itemize}

\section{Materials and methods}\label{sec:methods}

\subsection{Datasets and image acquisition}\label{sec:data}
\begin{figure}[ht]
\centering
\includegraphics[width=0.90\textwidth]{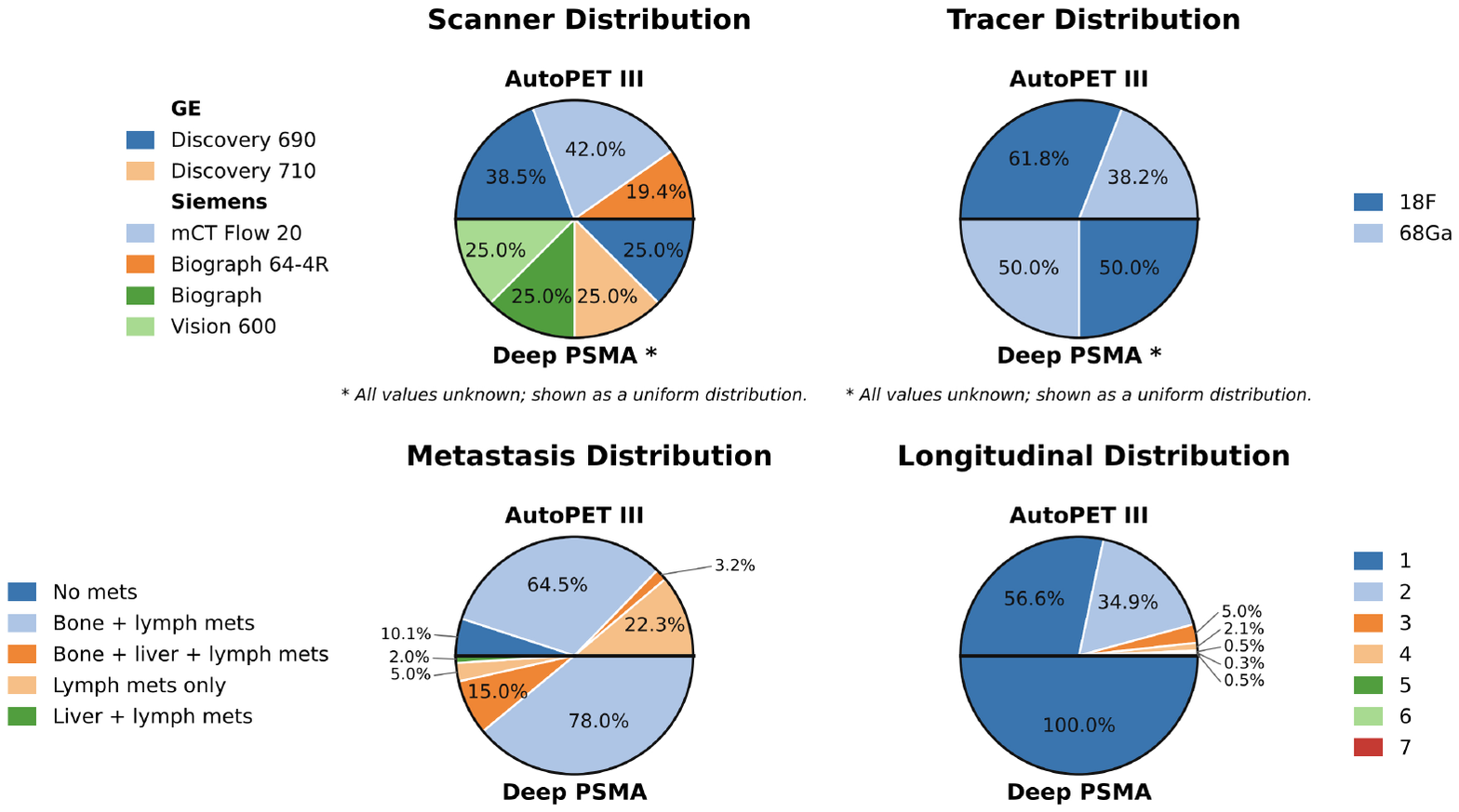}
\caption{Distribution of scanner, tracer, metastatic pattern, and number of scans per patient (longitudinal)
in the two evaluation cohorts. Each pie is divided into two halves, AutoPET~III (top) and Deep-PSMA (bottom),
each subdivided by that cohort's category proportions. Scanner and tracer are not reported per case for
Deep-PSMA ($^{*}$) and are shown as a uniform distribution over the reported categories.}\label{fig:data_dist}
\end{figure}

Two PSMA PET/CT datasets with different image acquisition and patient demographics were used (Table~\ref{tab:data_summary}, Fig.~\ref{fig:data_dist}).

\textbf{AutoPET III (PSMA)} is a publicly available cohort of 597 whole-body PSMA PET/CT scans with voxel-wise manual lesion annotations \citep{Jeblick26}. These scans were acquired with \textsuperscript{18}F-PSMA-1007 or \textsuperscript{68}Ga-PSMA-11 on three scanners (GE Discovery 690, Siemens Biograph mCT Flow 20, and Siemens Biograph 64-4R TruePoint) \citep{FEEDS} and cover the skull base to the mid-thigh. Cohort descriptions and statistics here follow our companion study \citep{FEEDS}. This cohort was partitioned into training, validation, and test sets (Section~\ref{sec:partition}) and was the only source of training data.

\textbf{Deep-PSMA} \citep{DeepPSMA25} is an independent public cohort of 100 patients with metastatic castration-resistant prostate cancer imaged before \textsuperscript{177}Lu-PSMA therapy. PSMA PET/CT was acquired with a mix
of \textsuperscript{68}Ga- and \textsuperscript{18}F-labeled tracers (\textsuperscript{68}Ga-PSMA and \textsuperscript{18}F-DCFPyL). Its reference segmentations were created by contouring at a fixed threshold of $\mathrm{SUV}\geq3$, after which an experienced nuclear medicine physician manually classified regions as malignant or physiological uptake. This protocol differs from the fully manual annotation of AutoPET~III. We used all 100 PSMA cases as an external test cohort.

For all cohorts, PET was used in SUV (body-weight) and CT was used on the PET grid. These two cohorts differ extensively in acquisition (in-plane resolution of 2.73--4.07~mm for AutoPET~III and 1.59--5.47~mm for Deep-PSMA), tracer, annotation procedure, lesion burden (mean 35.0 and 68.4 lesions per case), and mean lesion PET intensity (7.48 and 5.64), as well as in scanner and metastatic pattern. This makes them a demanding test for generalization. All results are reported on disease-positive cases.

\begin{table}[ht]
\caption{Summary of the two PSMA PET/CT cohorts. Pr = prostate cancer, Neg = negative, $dz$ = slice spacing, $\mu_{\mathrm{PET}}$ = mean PET intensity (per lesion). Cohort statistics are reproduced from the dataset summary in \citep{FEEDS}. The train/val/test counts sum to 593 rather than 597 since repeat scans of patients already in the training partition were excluded (Section~\ref{sec:partition}).}\label{tab:data_summary}
\begin{tabular*}{\linewidth}{@{\extracolsep{\fill}}lcc@{}}
\toprule
Property & AutoPET-III & Deep-PSMA \\
\midrule
\# Scans & 597 & 100 \\
Sex (M/F) & 597/0 & 100/0 \\
Disease (Pr/Neg) & 537/60 & 100/0 \\
\# Lesions per case & $35.04\pm51.70$ & $68.41\pm58.35$ \\
$\mu_{\mathrm{PET}}$ per lesion & $7.48\pm7.12$ & $5.64\pm3.33$\\
\# Slices per volume & 135--963 & 195--1261 \\
In-plane resolution (mm) & 2.73--4.07 & 1.59--5.47 \\
$dz$ between slices (mm) & 2.00--5.00 & 1.50--5.00\\
Train / val / test & 375/96/122 & 0/0/100 \\
\bottomrule
\end{tabular*}
\end{table}

\subsection{Data partitioning}\label{sec:partition}
AutoPET~III PSMA scans were partitioned once, before any modeling, into disjoint training (375 scans), validation (96), and test (122) sets, after excluding repeat scans of patients already represented in the training partition. The atlas and all networks were estimated from the training partition only. The validation set was used to select the loss configurations (which atlas maps weight which loss term) and the voting rule of the ensemble, and the test partition was never used for any design or tuning decision. The Deep-PSMA cohort serves as an external test cohort to assess generalizability across scan and annotation regimes. It is evaluated with the configuration frozen on the validation set and with no recalibration. All models use the final-epoch nnU-Net checkpoint for validation and testing (the nnU-Net default), with no checkpoint selection.

\subsection{Image pre-processing}
Pre-processing follows nnU-Net's self-configuring pipeline \citep{Isensee21}, using the plan configured for the AutoPET~III pan-cancer (FDG and PSMA) training data. Only the plan and the normalization constants were taken from this pan-cancer training set; none of its images were used to train the models, and the PSMA validation and test scans used here were not part of it. Each case consists of two input channels, CT (channel~0) and PET in units of SUV (channel~1). Volumes were cropped to the non-zero region and resampled to the plan's target spacing, the median of the training set, $3.0\times2.04\times2.04$~mm (axial slice thickness $\times$ in plane), using third-order spline interpolation for the images and first-order interpolation for the segmentation masks. The median volume size at this spacing is $300\times400\times400$ voxels. Both channels were normalized with nnU-Net's \texttt{CTNormalization}: intensities were clipped to the 0.5th and 99.5th percentiles of the foreground (lesion) voxels of the pan-cancer training set (CT: $[-832, 1128]$~HU; PET: $[1.04, 51.2]$~SUV) and then standardized with the mean and standard deviation of those same foreground voxels (CT: $107.7\pm286.3$; PET: $7.06\pm7.96$). These normalization statistics are fixed training-set constants, so no test-time information enters the normalization, and the same constants and spacing were applied to the external cohorts. The network is trained on patches of $112\times160\times128$ voxels (approximately $336\times326\times261$~mm), with a batch size of 2 and batch Dice enabled, as selected by the planner for the pan-cancer data. The reference lesion masks were converted to binary masks. In addition, the atlas of each training patient was brought to the same preprocessed frame as the image by applying the same axis transpose, the same cropping bounding box, and a proportional rescale to the resampled shape (Section~\ref{sec:trainimpl}).

\subsection{Population atlas construction}\label{sec:atlas}
\begin{figure}[ht]
\centering
\includegraphics[width=0.99\textwidth]{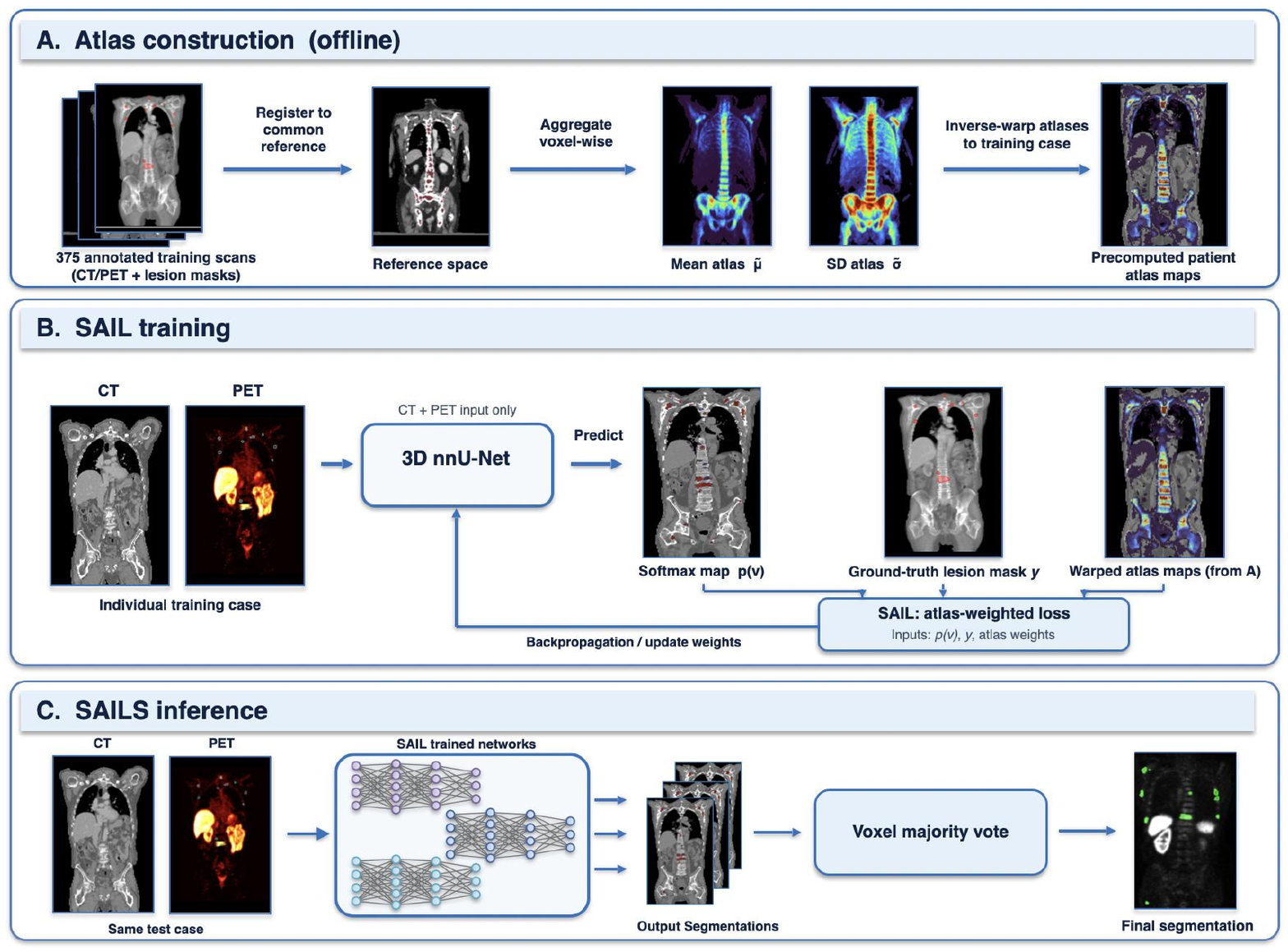}
\caption{Overview of the proposed SAIL and SAILS framework.
\textbf{(A)} Atlas construction: 375 annotated PSMA PET/CT training scans are registered to a common reference, and their warped lesion masks are aggregated voxel-wise to construct the mean ($\mu$) and standard deviation ($\sigma$) metastasis atlases. These atlases are then mapped to individual training cases.
\textbf{(B)} SAIL training: the atlas maps, reference lesion mask, and network predictions are used to compute an atlas-weighted loss that modulates cross-entropy according to spatial metastasis statistics and prediction confidence. The segmentation network receives only CT and PET as inputs.
\textbf{(C) }SAILS inference: networks trained with different atlas-weighted loss configurations produce binary segmentations that are combined through voxel-wise majority voting. No atlas or registration is required at inference.}
\label{fig:overview}
\end{figure}

Spatial population atlases summarize where a structure or disease occurs across a cohort by registering every subject to a common reference frame and aggregating their labels voxel-wise. Following the strategy used to build whole-body atlases from UK Biobank MRI \citep{Starck24}, we construct a spatial atlas of PSMA-avid metastases from the AutoPET III training partition.
The training subject with median body volume was selected as the registration reference, to minimize average deformation. Every training CT was registered to the reference by affine initialization followed by symmetric diffeomorphic (SyN) registration with a cross-correlation metric \citep{Avants08} using ANTsPy with a grad step of 0.2 and registration iterations of (80, 50, 20). Registering all training scans took 19 hours on 32 CPU cores with 196 GB of memory. The resulting transform was applied to the binary lesion mask with nearest-neighbor interpolation. The voxel-wise mean over the $N{=}375$ training scans (including disease-negative scans, whose lesion masks are empty) gives the mean atlas $\mu(v)$, the empirical marginal probability of a lesion at $v$. Lesions occupy a small fraction of whole-body voxels, so $\mu$ is small everywhere. Its maximum value across any voxel is $0.1604$. In addition to the mean, a per-voxel variance map $\sigma^2(v)$ was computed as $\sigma^2(v)=\frac{1}{N}\sum_i(\ell_i(v)-\mu(v))^2$ over the warped training masks $\ell_i$ (Fig.~\ref{fig:overview}). As the lesion masks are binary, $\sigma^2(v)=\mu(v)\,(1-\mu(v))$, so the standard deviation map is a monotone, square-root-like transform of the mean atlas that gives relatively more weight to sites where metastases occur but are infrequent

To use the atlas for training, the atlas is mapped to each patient's anatomy through the inverse of that patient's registration transform with linear interpolation (nearest-neighbor interpolation would inject plateaus). This mapping is computed once offline and stored for each training patient. Both maps are normalized to $[0, 1]$ with fixed population constants as follows (Eqs.~\eqref{eq:mutilde} and~\eqref{eq:sigtilde}):
\begin{align}
\tilde\mu(v) &= \mathrm{clip}\!\left(\frac{\mu(v)}{\mu_{\max}},0,1\right), \quad \mu_{\max} = \text{maximum voxel value in $\mu(v)$}\label{eq:mutilde}\\
\tilde\sigma(v) &= \mathrm{clip}\!\left(\frac{\sqrt{\sigma^2(v)}}{\sqrt{\sigma^2(v)}_{\max}},0,1\right), \sqrt{\sigma^2(v)}_{\max} = \text{maximum voxel value in $\sqrt{\sigma^2(v)}$}.\label{eq:sigtilde}
\end{align}
Each map is divided by its own maximum, so the clip serves only as a numerical safeguard. The normalized weight reaches 1 only at the voxels where the map attains its maximum. We use the standard deviation $\sqrt{\sigma^2}$ rather than the variance $\sigma^2$ because it is expressed in the same units as the atlas probability $\mu$ (variance carries units of probability squared). Furthermore, the square root is a concave transform that stretches the small values found over most of the atlas support. The variance would compress them toward zero and make the weight negligible nearly everywhere so that only the few highest-variance voxels carry meaningful signal. As a result, $\tilde\sigma$ assigns substantial weight to rare but clinically important sites such as the liver, where $\tilde\mu$ is close to zero.

\subsection{Atlas-weighted loss}
\textbf{Notation.} The loss is computed on training patches, the sub-volumes of $112\times160\times128$ voxels that nnU-Net samples from the training scans at each iteration. Let $\Omega$ denote the set of voxels in a patch. At each voxel $v \in \Omega$, let $p_v$ be the network's softmax foreground probability (the predicted lesion probability) and $y_v \in \{0,1\}$ the corresponding ground-truth label (1 for lesion). We call voxels with $y_v=1$ foreground (FG) voxels, collected in $\Omega_{\text{FG}} = \{v \in \Omega : y_v = 1\}$, which are the reference lesion voxels. Voxels with $y_v=0$ are background (BG) voxels, collected in $\Omega_{\text{BG}} = \{v \in \Omega : y_v = 0\}$.

\textbf{Standard loss.} The standard nnU-Net training objective combines cross-entropy (CE) loss and soft Dice loss to jointly optimize voxel-wise classification and spatial overlap:

$$
\mathcal{L}_{\mathrm{nnU-Net}}=\mathcal{L}_{\mathrm{CE}}+\mathcal{L}_{\mathrm{Dice}}.
$$
The soft Dice loss is computed over all voxels $\Omega_B$ of a training batch (batch Dice),

\begin{equation}
\mathcal{L}_{\mathrm{Dice}}=1-\frac{2\sum_{v\in\Omega_B}p_v\,y_v}{\sum_{v\in\Omega_B}p_v+\sum_{v\in\Omega_B}y_v},
\label{eq:dice}
\end{equation}

and is therefore a single ratio of sums rather than a sum of per-voxel terms. The CE loss, in contrast, is a sum of per-voxel terms.

The CE loss can be split into foreground (FG) and background (BG) terms:

\begin{equation}
\mathcal{L}_{\mathrm{CE}} =
-\frac{1}{|\Omega|}
\left[
\underbrace{\sum_{v\in\Omega_{\mathrm{FG}}}\log(p_v)}_{\mathrm{FG}}
+
\underbrace{\sum_{v\in\Omega_{\mathrm{BG}}}\log(1-p_v)}_{\mathrm{BG}}
\right]
\label{eq:ce_std}
\end{equation}

The foreground term penalizes low predicted probabilities for true lesion voxels (missed lesion voxels). The background term penalizes high foreground probabilities assigned to background voxels (false positive voxels).

\textbf{Idea.} The standard CE loss assigns equal importance to all voxels within a patch, without accounting for the spatial statistics of metastasis occurrence. The atlas-weighted loss is designed to roughly mimic how a clinician weighs errors: a missed lesion should cost more where metastases are known to occur, including high-risk sites (e.g., bones, liver), even if the probability of occurrence is low (e.g. liver). A false positive should cost more or less depending on how plausible a metastasis is at that site, and this extra penalty should apply only while the network is still wrong or uncertain at that voxel. The population-level atlas constructed in Sect.~\ref{sec:atlas} provides voxel-wise statistics of metastasis occurrence through the estimated mean and standard deviation. We leverage these statistics to adaptively modulate the contributions of foreground (FG) and background (BG) voxels to the segmentation objective, using two ingredients: an atlas weight that depends on location, and a confidence kernel that depends on the current prediction.

\textbf{Atlas weights.} The atlas weights are $A_{\mathrm{fg}}(v)\in\{\tilde{\mu}(v),\tilde{\sigma}(v)\}$ for FG voxels and $A_{\mathrm{bg}}(v)\in\{\tilde{\sigma}(v),1-\tilde{\mu}(v)\}$ for BG voxels, where $\tilde{\mu}$ and $\tilde{\sigma}$ denote the normalized atlas mean and standard deviation (Eqs.~\eqref{eq:mutilde} and~\eqref{eq:sigtilde}). All of them lie in $[0,1]$. The different atlas weights signify the following:
\begin{itemize}
    \item $A_{\mathrm{fg}}=\tilde\mu$: missed lesions cost more where metastases are most frequent.
    \item $A_{\mathrm{fg}}=\tilde\sigma$: missed lesions cost more wherever metastases occur variably, which spreads emphasis to less frequent but clinically important sites.
    \item $A_{\mathrm{bg}}=1-\tilde\mu$: false positives cost more where metastases are rarely observed.
    \item $A_{\mathrm{bg}}=\tilde\sigma$: false positives cost more at sites where metastases do occur, so the network cannot rely on location alone there and must separate lesions from physiological or benign uptake using image features.

\end{itemize}

\textbf{Confidence kernel.} Rather than directly modifying the predicted voxel-wise probabilities using atlas statistics, we adopt an exponential weighting kernel $g_\alpha(q)$, motivated by \cite{dong2021recognition,Lin17Focal,Abraham19FocalTversky}:

\begin{equation}
g_\alpha(q)=
\frac{e^{-\alpha q}-e^{-\alpha}}
{1-e^{-\alpha}},
\qquad q\in[0,1],\ \alpha>0.
\label{eq:kernel}
\end{equation}

\begin{figure}[ht]
\centering
\includegraphics[width=0.9\textwidth]{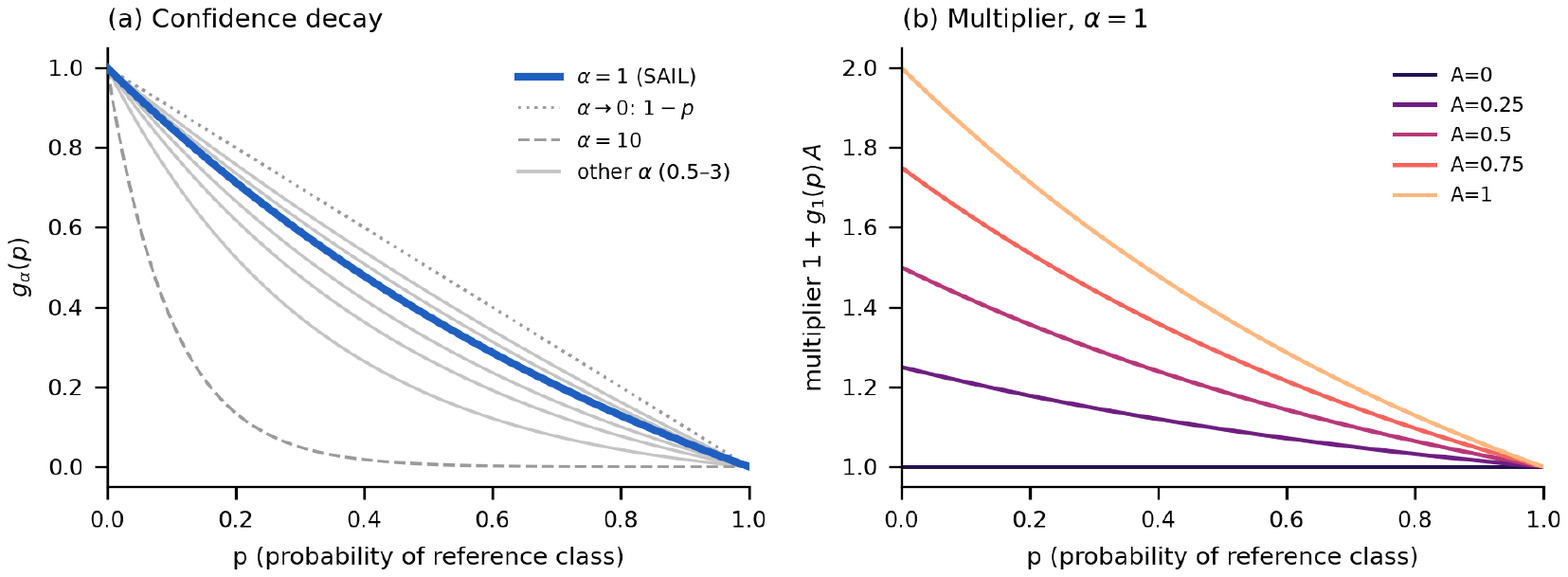}
\caption{Behavior of the confidence kernel and the atlas-weighted multiplier. (a) The confidence kernel $g_\alpha(p)$ for several $\alpha$; dotted line is the $\alpha\to0$ limit $1-p$ and the dashed line is $\alpha=10$, close to the step-function limit. All experiments use $\alpha=1$. (b) The resulting multiplier $1+g_1(p)A$ for several atlas weights $A$ (with $\alpha=1$); $A=0$ recovers standard cross-entropy. For background voxels, $p$ is replaced by $1-p$.}\label{fig:decay}
\end{figure}

For FG voxels, we set $q=p_v$, the predicted lesion probability at voxel $v$, such that the kernel assigns larger weights to incorrect predictions with low foreground probabilities and progressively decreases toward zero as $p_v$ approaches one. Conversely, for BG voxels, we set $q=1-p_v$, thereby emphasizing incorrectly predicted background voxels with high foreground probabilities while reducing the weight for correctly classified background voxels. The parameter $\alpha$ controls the shape of the decay, from nearly linear for small $\alpha$ to nearly a step function for large $\alpha$ (Fig.~\ref{fig:decay}a). We fixed $\alpha=1$ for all experiments.

\textbf{Atlas-weighted cross-entropy.} Combining the two ingredients gives the foreground and background weighting factors
\begin{equation}
W_v^{\mathrm{fg}}=g_\alpha(p_v)A_{\mathrm{fg}}(v),
\qquad
W_v^{\mathrm{bg}}=g_\alpha(1-p_v)A_{\mathrm{bg}}(v),
\label{eq:w}
\end{equation}
which multiply the corresponding terms of Eq.~\eqref{eq:ce_std}. The resulting atlas-weighted cross-entropy loss is formulated as

\begin{equation}
\mathcal{L}_{\mathrm{CE}}^{\mathrm{AW}}
=
-\frac{1}{|\Omega|}
\left[
\underbrace{
\sum_{v\in\Omega_{\mathrm{FG}}}
\left(1+W_v^{\mathrm{fg}}\right)\log(p_v)
}_{\mathrm{FG}}
+
\underbrace{
\sum_{v\in\Omega_{\mathrm{BG}}}
\left(1+W_v^{\mathrm{bg}}\right)\log(1-p_v)
}_{\mathrm{BG}}
\right].\label{eq:ce}
\end{equation}

As shown in Table \ref{tab:atlas_weight_behavior}, for correctly predicted FG voxels ($p_v\to 1$), the exponential kernel $g_\alpha(p_v)$ approaches zero, suppressing the contribution of the atlas-derived weight $A_{\mathrm{fg}}(v)$. Similarly, for correctly predicted BG voxels ($p_v\to 0$), the kernel $g_\alpha(1-p_v)$ approaches zero, minimizing the influence of $A_{\mathrm{bg}}(v)$. Consequently, the proposed weighting strategy adaptively suppresses the influence of atlas statistics for correctly classified voxels. Conversely, for confidently incorrect FG predictions ($p_v\to 0$), $g_\alpha(p_v)$ approaches unity, maximizing the atlas-derived weighting contribution and increasing the penalty for misclassification. The same behavior applies to confidently incorrect BG predictions ($p_v\to 1$), where $g_\alpha(1-p_v)$ approaches unity. Therefore, the proposed loss emphasizes difficult voxel predictions with population-level anatomical priors. Each multiplier $1+W_v$ lies in $[1,2]$ (Fig.~\ref{fig:decay}b) and equals 1, recovering the standard CE, wherever the atlas weight is zero. The atlas therefore adds emphasis but never reduces the weight of any voxel. Either weight can be switched off ($W\equiv0$), which gives foreground-only and background-only variants. Finally, we combine our atlas-weighted loss with the standard soft  Dice loss (Eq.~\eqref{eq:dice}) to obtain the full training objective, which we call SAIL (\textbf{S}patial \textbf{A}tlas-\textbf{I}nformed \textbf{L}oss):

\begin{equation}
\mathcal{L}_{\mathrm{SAIL}}=\lambda_{\mathrm{CE}}\,\mathcal{L}_{\mathrm{CE}}^{\mathrm{AW}}+\lambda_{\mathrm{Dice}}\,\mathcal{L}_{\mathrm{Dice}},
\label{eq:awl}
\end{equation}
with $\lambda_{\mathrm{CE}}=\lambda_{\mathrm{Dice}}=1$ and $\mathcal{L}_{\mathrm{Dice}}$ the standard nnU-Net batch soft-Dice loss excluding background. Only the CE term is modified, CE is a sum of per-voxel terms, so a per-voxel weight is well defined, whereas soft Dice (Eq.~\eqref{eq:dice}) is a single ratio over the whole batch with no per-voxel term to weight. The loss is applied at every deep-supervision output with nnU-Net's halving scheme, with the atlas weight resampled to each output resolution by trilinear interpolation. For numerical stability, $p_v$ is clamped to $[10^{-7},1-10^{-7}]$ when evaluating $\log p_v$ and $\log(1-p_v)$.

\begin{table}[t]
\centering
\caption{Effect of the exponential weighting kernel on
atlas-derived contributions for correctly and incorrectly
predicted foreground (FG, reference lesion) and background (BG) voxels.}
\label{tab:loss_table}\label{tab:atlas_weight_behavior}
\renewcommand{\arraystretch}{1.15}
\setlength{\tabcolsep}{4pt}
\small
\begin{tabular}{c c c c c}
\toprule
\textbf{Class} &
\textbf{Prediction} &
$\boldsymbol{p_v}$ &
\textbf{Kernel} &
\textbf{CE Weight ($1+W_v$)} \\
\midrule

\multirow{2}{*}{FG}
& Correct
& $\to 1$
& $g_\alpha(p_v)\to 0$
& $\to 1$ \\

& Incorrect
& $\to 0$
& $g_\alpha(p_v)\to 1$
& $\to 1+A_{\mathrm{fg}}(v)$ \\

\midrule

\multirow{2}{*}{BG}
& Correct
& $\to 0$
& $g_\alpha(1-p_v)\to 0$
& $\to 1$ \\

& Incorrect
& $\to 1$
& $g_\alpha(1-p_v)\to 1$
& $\to 1+A_{\mathrm{bg}}(v)$ \\

\bottomrule
\end{tabular}
\end{table}

\subsection{Network, training and implementation}\label{sec:trainimpl}
All models are 3D full-resolution nnU-Net \citep{Isensee21} networks trained from scratch on the training partition with the framework's default self-configured hyperparameters (SGD with Nesterov momentum 0.99, initial learning rate 0.01 with polynomial decay, deep supervision, standard spatial and intensity augmentation), except for the 300 epoch schedule (default nnU-Net uses 1,000). All configurations share the same data, partition, random seed, augmentation, patch size ($112 \times 160 \times 128$ voxels), batch size (2), and 300-epoch schedule. For the atlas-weighted configurations, the atlas channels are appended to the image tensor at data-loading time only so that they pass through exactly the same spatial augmentation (rotation, scaling, mirroring) and remain aligned with the image patch. They are separated from the image channels before the forward pass, so the network receives only the PET and CT channels.

The atlas of each patient is aligned to the preprocessed image by applying the same axis transpose as the nnU-Net's preprocessor, applying the same cropping bounding box recorded in the case properties, and applying a proportional resize to the resampled shape. At validation and inference, the atlas channels are removed and no atlas is loaded. The prediction of the trained network is therefore identical in form and cost to the baseline.

\subsection{Compared methods and ablation studies}\label{sec:ablation}

We compare the proposed loss against the baseline and against other ways of bringing the same atlas into an nnU-Net, grouped by where the atlas enters the pipeline (Table~\ref{tab:configs}). All trained models share the same data, partition, seed, augmentation, and 300-epoch schedule, and differ only in the component listed.

\textbf{(A) Baseline.} A standard nnU-Net trained with the default cross-entropy plus soft Dice loss with PET and CT images as inputs, with no atlas information.

\textbf{(B) Atlas as input.} The warped mean atlas $\tilde\mu$ is appended as a third input channel alongside CT and PET, and the network is retrained with the standard loss. The atlas undergoes the same spatial augmentation as the images and is excluded from intensity augmentation. This variant lets the network learn how to use the prior, but, unlike the proposed loss, it requires registering the atlas to every patient at inference. It was evaluated on the validation set only.

\textbf{(C) Inference-time fusion.} The baseline network is left unchanged and its softmax output is combined with the warped mean atlas at test time. In calibrated log-odds fusion \citep{ChauAtlasFusion26}, the atlas log-odds are added to the network log-odds after calibration, so the atlas has no effect outside its support. Its parameters were tuned on the validation set, as in \citep{ChauAtlasFusion26}.

\textbf{(D) Atlas-weighted loss (SAIL, ours).} The atlas enters only the training loss (Eq.~\eqref{eq:ce}), with $\alpha=1$. We vary which term is weighted (foreground only, background only, or both) and by which atlas map ($\tilde\mu$, $\tilde\sigma$, or $1-\tilde\mu$), giving seven configurations (Table~\ref{tab:configs}, D1--D7):
\begin{itemize}
    \item \emph{Foreground only} (missed lesions): D1 $A_{\mathrm{fg}}=\tilde\mu$; D2 $A_{\mathrm{fg}}=\tilde\sigma$.
    \item \emph{Background only} (false positives): D3 $A_{\mathrm{bg}}=1-\tilde\mu$.
    \item \emph{Both terms} (foreground / background): D4 $\tilde\mu$ / $1-\tilde\mu$; D5 $\tilde\sigma$ / $\tilde\sigma$; D6 $\tilde\sigma$ / $1-\tilde\mu$; D7 $\tilde\mu$ / $\tilde\sigma$.
\end{itemize}

\begin{table}[htbp]\footnotesize\setlength{\tabcolsep}{3pt}
\caption{Configurations compared, grouped by where the atlas enters the pipeline. Atlas: which atlas statistic is used (none, mean $\tilde\mu$, standard deviation $\tilde\sigma$, or both). $A_{\mathrm{fg}}$, $A_{\mathrm{bg}}$: atlas maps weighting the foreground (missed lesion) and background (false-positive) cross-entropy terms (Eq.~\eqref{eq:w}); a dash means that term is unweighted. All loss configurations use $\alpha=1$. Atlas at test: whether per-patient registration of the atlas is needed at test time. SAILS combines the predictions of D3, D5 and D6 by majority voting (Section~\ref{sec:ablation}).}\label{tab:configs}
\begin{tabularx}{\linewidth}{@{}lXcccc@{}}
\toprule
Group & Configuration & Atlas & $A_{\mathrm{fg}}$ & $A_{\mathrm{bg}}$ & Atlas at test \\
\midrule
(A) Baseline & nnU-Net & None & --- & --- & No \\
\midrule
(B) Atlas as input & Extra input channel & Mean & --- & --- & Yes \\
\midrule
(C) Inference fusion & Calibrated log-odds \citep{ChauAtlasFusion26} & Mean & --- & --- & Yes \\
\midrule
(D) Atlas-weighted & D1 Fg only & Mean & $\tilde\mu$ & --- & No \\
\quad loss & D2 Fg only & Std & $\tilde\sigma$ & --- & No \\
 & D3 Bg only & Mean & --- & $1-\tilde\mu$ & No \\
 & D4 Fg + Bg & Mean & $\tilde\mu$ & $1-\tilde\mu$ & No \\
 & D5 Fg + Bg & Std & $\tilde\sigma$ & $\tilde\sigma$ & No \\
 & D6 Fg + Bg & Both & $\tilde\sigma$ & $1-\tilde\mu$ & No \\
 & D7 Fg + Bg & Both & $\tilde\mu$ & $\tilde\sigma$ & No \\
\midrule
(E) SAILS & Vote of D3, D5, D6 & Both & --- & --- & No \\
\bottomrule
\end{tabularx}
\end{table}

\textbf{(E) SAILS: majority voting of atlas-weighted networks (ours).} Each atlas-weighted configuration trades missed lesions against false positives differently. We therefore combine the binary predictions of three configurations without retraining. The members are the configuration with the highest validation Dice (D5) and the two configurations that increased validation Dice while reducing both false-negative and false-positive volume (D3 and D6). We compare two voting rules with $k\in\{1,2,3\}$. In the voxel vote, a voxel is labeled lesion if at least $k$ members predict it. In the lesion vote, the union of the member predictions is split into connected components (18-connectivity), and a component is kept if at least $k$ members predict any voxel inside it. With $k=1$ both rules reduce to the union of the member predictions. The voting rule is selected on the validation set with the same criterion as the members: it must increase Dice and reduce both false-negative and false-positive volume. The selected rule is referred to as SAILS (\textbf{S}patial \textbf{A}tlas-\textbf{I}nformed \textbf{L}oss en\textbf{S}emble). Each member is a standard nnU-Net, so SAILS needs no atlas or registration at inference.

All configurations and voting variants are compared on the validation set only. SAILS is evaluated on the test cohorts.

\subsection{Evaluation Methods}\label{sec:eval}
All models were evaluated on the held-out AutoPET~III PSMA test set and on the external Deep-PSMA cohort using the same predictions, post-processing and metrics. Metrics were computed at three levels (voxel, lesion, and anatomic region). Unless stated otherwise, all metrics are computed on disease-positive cases only, and predictions are the $\arg\max$ of the network's softmax output with no threshold tuning.

\textbf{Voxel level:} Voxel-level evaluation captures overall volumetric overlap between predicted and reference lesion masks, independent of how many discrete lesions are involved. We compute the Dice similarity coefficient between the predicted and reference whole-body lesion masks for every case and report the mean $\pm$ SD over cases. We also report the false-positive volume (FPVol, the total volume of spuriously detected lesion components that do not overlap any reference lesion) and false-negative volume (FNVol, the total volume of reference lesions that are missed entirely), in cm$^3$ (cc), as per-case means, following \citep{Rokuss24,FEEDS}. A prediction with zero overlap with the reference scores a Dice of 0 rather than being excluded, so that missed cases are penalized instead of dropped from the average. FNVol therefore counts only lesions that are missed completely, and FPVol only detections that are completely spurious. Reporting them alongside Dice shows which kind of error a change in Dice comes from: for example, Dice can stay similar while FNVol falls and FPVol rises, if newly found lesions are offset by new false positives.

\textbf{Lesion-level evaluation:} Dice can improve either because an already-detected lesion is segmented more completely or if a previously missed lesion is found. Only the latter changes how many disease sites are identified. We therefore also evaluate how many discrete lesions are detected, regardless of how accurately their boundaries are drawn. This distinction is clinically relevant: in oligometastatic prostate cancer, where disease is confined to a few sites, metastasis-directed radiotherapy has been associated with improved outcomes \citep{alberto2022role,phillips2020outcomes}, so the number of lesions found can matter as much as how well each is outlined.

Reference and predicted masks were split into discrete lesions by 3D connected-component analysis (18-connectivity). A reference lesion counted as detected if any predicted voxel overlapped it, and a predicted component that overlapped no reference lesion counted as a false positive. We report lesion-level sensitivity (detection rate) and positive predictive value (PPV), both pooled over all lesions in a cohort before the ratio is taken, so patients with many lesions are not down-weighted relative to those with few. This matters here because mean lesion burden differs almost sevenfold across cohorts (Table~\ref{tab:data_summary}). We also report the total number of additional reference lesions detected, and of additional false-positive components, relative to the baseline in each cohort. For the statistical test only, sensitivity is also computed separately for each patient, because the Wilcoxon test requires one paired value per patient. To characterize the additional false positives, we identified the false-positive components of SAILS that did not overlap any false-positive component of the baseline and recorded their volume and maximum SUV (SUV$_{\max}$).

\textbf{Anatomic-region level:} Not all missed or spurious lesions carry equal clinical weight. For example, a missed liver metastasis or a missed high-risk bone lesion can change a patient's management in ways that a missed small lymph node metastasis would not. We therefore additionally stratified performance in three regions chosen for their clinical consequence in prostate cancer. Bone is the dominant site of metastasis \citep{Gandaglia14}. Liver and lung involvement is less frequent but is associated with worse survival \citep{Kadeerhan23}. High-risk bone lesions, defined below, are those for which prophylactic radiotherapy has been evaluated to prevent skeletal-related events \citep{Gillespie24HighRiskBone}, so missed lesions there can change management.

Liver and bone were segmented on the CT scans using the pretrained TotalSegmentator model \citep{Wasserthal23TotalSegmentator}, and Dice, FNVol, and FPVol were recomputed restricted to voxels within each structure. These region-level values describe the metastases within each region (for example, metastasis Dice within bone), not the segmentation of the organ itself. Following our prior study \citep{FEEDS}, region-level metrics were computed per TotalSegmentator label instance and included only instances with at least 0.4cc of reference lesion overlap. To compare methods on the same cases, region-level means are computed over cases evaluable for all compared methods.

We additionally evaluated \textbf{high-risk bone} metastases, following the criteria from a multicenter randomized trial of prophylactic radiotherapy \citep{Gillespie24HighRiskBone}, also used in our companion study \citep{FEEDS}. A bone metastasis is high-risk if it meets at least one of the following: bulky ($\geq2$~cm) disease; involvement of the hip, shoulder or sacroiliac joint; disease in long bones occupying up to two-thirds of cortical thickness; or disease in the junctional spine (C7--T1, T12--L1, L5--S1) or with posterior element involvement \citep{Gillespie24HighRiskBone}. These criteria mark bone lesions where symptomatic progression or skeletal-related events are more likely, and where segmentation errors are therefore more consequential for treatment planning than in bone disease generally.

\textbf{Statistics.} All comparisons are paired against the baseline. For each metric we report the mean difference (the mean over patients of each patient's difference, model minus baseline; for example, $\Delta$FPVol is the average of the per-patient FPVol differences) with a bootstrap 95\% confidence interval (CI). On the validation set, configurations and voting variants are compared by these differences only, without significance testing. On the test cohorts, SAILS is compared with the baseline using the two-sided Wilcoxon signed-rank test, with $p<0.01$ considered significant; $p$ values for Dice and false-negative volume, the primary and key secondary metrics, are Holm-adjusted. Inference-time fusion is reported descriptively. Region-level comparisons use per-case Dice within each region. The primary comparison metric is the paired mean Dice difference to the baseline on the AutoPET~III test set. As SAIL is designed to reduce missed lesions, false-negative volume and pooled lesion sensitivity are the key secondary metrics, and we also report false-positive volume.

\section{Results}\label{sec:results}

\subsection{Ablation and model selection on the validation set}\label{sec:res_val}
Table~\ref{tab:val} lists validation results for every configuration and voting variant. Among the single atlas-weighted configurations, only foreground weighting by the mean atlas lowered Dice with a 95\% CI below zero (D1, $-0.013$, 95\% CI $-0.025$ to $-0.000$); the other six changed Dice by $-0.001$ to $+0.005$. Only D3 and D6 increased Dice while reducing both false-negative and false-positive volume. D3, D5 and D6 were therefore combined by voting.

\textbf{Voting.} The union of the three predictions ($k=1$) raised sensitivity to 0.892 but increased false-positive volume by 12.7~cc and lowered Dice. Requiring agreement of all three networks ($k=3$) reduced false-positive volume the most ($-5.9$~cc for the voxel vote) but increased false-negative volume. The voxel vote with $k=2$ was the only variant that increased Dice ($+0.012$, 95\% CI $+0.002$ to $+0.022$) while reducing both false-negative ($-2.2$~cc) and false-positive volume ($-1.4$~cc), and it was selected as SAILS. The lesion vote keeps whole components of the union, so it retained more false-positive volume than the voxel vote at the same $k$.

\textbf{Atlas source.}
Weighting the foreground term only by the mean atlas $\tilde\mu$ was the weakest single-map choice on the validation set. It was the only configuration that lowered Dice (--0.013). Replacing $\tilde \mu$ with $\tilde \sigma$ moved validation Dice to $+0.001$, consistent with $\tilde\sigma$ spreading weight beyond the few voxels of peak prevalence (Section~\ref{sec:atlas}).

\textbf{Background weighting.} On the validation set, background-only weighting reduced false-positive volume slightly ($-0.7$~cc), and only one of the four combined configurations lowered it (D6, $-2.7$~cc, 95\% CI $-9.2$ to $+1.8$), whereas D4, D5 and D7 raised it by $+3.0$ to $+8.9$~cc. With $A_{\mathrm{bg}}=1-\tilde\mu$ the weight is close to one almost everywhere, so it acts as a near-uniform up-weighting of background errors outside the atlas support and relatively less up-weighting inside it. This shifts the decision boundary toward foreground at atlas-supported voxels. Consistent with this, background-only weighting gave the largest reduction in false-negative volume on the validation set (Table~\ref{tab:val}).

\textbf{Atlas as input.} Supplying the mean atlas $\tilde\mu$ as a third input channel (B) did not improve on the seeded 300-epoch baseline on the validation set, so we did not evaluate it on the test sets. Whole-body Dice was 0.508 vs. 0.519 (mean) and 0.597 vs. 0.600 (median). FN volume was unchanged (13.42 vs. 13.42 cc), FP volume was slightly higher ($18.50$ vs.\ $18.20$ cc), and pooled lesion sensitivity was $0.861$ vs.\ $0.859$ (PPV $0.772$ vs.\ $0.784$). The paired Dice difference was $-0.011$ (95\% CI $-0.025$ to $+0.002$), so (B) was not non-inferior at the 0.02 margin.

\subsection{Voxel-level performance on the test cohorts}

\textbf{AutoPET~III.} SAILS improved whole-body Dice on the held-out test set from 0.520 to 0.531 ($+0.011$, 95\% CI $+0.000$ to $+0.022$; $p=0.006$; Table~\ref{tab:voxel}). False-negative volume decreased from 12.1 to 9.5~cc per case ($p<0.01$), compared with 11.0~cc for inference-time fusion. False-positive volume increased from 5.5 to 7.8~cc per case ($p<0.001$), close to that of fusion (7.1~cc).

\textbf{Deep-PSMA.} On the external cohort, SAILS did not change whole-body Dice significantly (0.582 vs.\ 0.590; $-0.009$, 95\% CI $-0.021$ to $+0.002$; $p=0.36$). False-negative volume decreased by 2.7~cc per case ($p<0.01$). False-positive volume increased from 25.2 to 43.4~cc per case ($p<0.001$). Inference-time fusion, which requires registering the atlas to every test patient, improved Dice to 0.619 and added little false-positive volume (27.1~cc).

\subsection{Lesion-level performance on the test cohorts}
The increase in detected lesions carried over to the external cohort (Table~\ref{tab:lesion}). On AutoPET~III, SAILS detected 46 more of the 3,231 reference lesions than the baseline, raising pooled sensitivity from 0.843 to 0.857 ($p=0.004$). On Deep-PSMA, the cohort with the highest lesion burden, it detected 126 more of the 6,841 reference lesions, raising pooled sensitivity from 0.751 to 0.769 ($p<0.001$).

SAILS added 150 false-positive components on AutoPET~III and 198 on Deep-PSMA, so pooled PPV decreased (0.793 to 0.759 and 0.887 to 0.865). Most of these components were small: their median volume was 0.13~cc on AutoPET~III and 0.54~cc on Deep-PSMA, and 92\% and 65\%, respectively, were smaller than 1~cc. On Deep-PSMA, their uptake was similar to that of the baseline's false positives (median SUV$_{\max}$ 9.0 vs.\ 8.9), indicating avid foci that were not labeled as metastases in the reference rather than sub-threshold noise. On AutoPET~III, they were fainter than the baseline's false positives (median SUV$_{\max}$ 4.1 vs.\ 5.1; 31\% vs.\ 12\% with SUV$_{\max}<3$).

\begin{table}[ht]
\caption{Validation-set results (AutoPET~III, 84 disease-positive cases). Fg/Bg: atlas map weighting the foreground/background cross-entropy term, as in Table~\ref{tab:configs}; all loss configurations use $\alpha=1$. Voting variants combine D3, D5 and D6: voxel vote, a voxel is labeled lesion if at least $k$ networks predict it; lesion vote, a connected component of the union is kept if at least $k$ networks detect it. $\Delta$Dice: paired mean difference to the baseline with bootstrap 95\% CI. $\Delta$FNVol / $\Delta$FPVol: paired mean change in false-negative / false-positive volume per case (cc); negative values are better. Inference-time fusion parameters were tuned on this set. $\dagger$: selected as SAILS.}\label{tab:val}
\small
\centering
\begin{tabular}{@{}lcccc@{}}
\toprule
Configuration & Dice & $\Delta$Dice (95\% CI) & $\Delta$FNVol & $\Delta$FPVol \\
\midrule
Baseline & 0.519 & --- & --- & --- \\
Atlas as input (B) & 0.508 & $-0.011$ [$-0.025$, $+0.002$] & $0.0$ & $+0.3$ \\
Inference-time fusion (C) & 0.526 & $+0.007$ [$+0.002$, $+0.013$] & $-1.0$ & $+4.2$ \\
\midrule
\multicolumn{5}{@{}l}{\textit{(D) Atlas-weighted loss configurations}} \\
D1: Fg~$\tilde\mu$ & 0.506 & $-0.013$ [$-0.025$, $-0.000$] & $-1.2$ & $-1.1$ \\
D2: Fg~$\tilde\sigma$ & 0.520 & $+0.001$ [$-0.010$, $+0.012$] & $-1.1$ & $+5.2$ \\
D3: Bg~$1-\tilde\mu$ & 0.523 & $+0.004$ [$-0.007$, $+0.016$] & $-2.6$ & $-0.7$ \\
D4: Fg~$\tilde\mu$ + Bg~$1-\tilde\mu$ & 0.518 & $-0.001$ [$-0.013$, $+0.010$] & $+0.4$ & $+3.0$ \\
D5: Fg~$\tilde\sigma$ + Bg~$\tilde\sigma$ & 0.524 & $+0.005$ [$-0.007$, $+0.017$] & $-0.7$ & $+8.9$ \\
D6: Fg~$\tilde\sigma$ + Bg~$1-\tilde\mu$ & 0.520 & $+0.001$ [$-0.010$, $+0.012$] & $-1.7$ & $-2.7$ \\
D7: Fg~$\tilde\mu$ + Bg~$\tilde\sigma$ & 0.519 & $-0.000$ [$-0.011$, $+0.012$] & $-1.5$ & $+3.9$ \\
\midrule
\multicolumn{5}{@{}l}{\textit{(E) Voting of D3, D5 and D6}} \\
Union ($k=1$) & 0.506 & $-0.013$ [$-0.027$, $+0.002$] & $-4.2$ & $+12.7$ \\
Voxel vote, $k=2$ $\dagger$ & \textbf{0.531} & $+0.012$ [$+0.002$, $+0.022$] & $-2.2$ & $-1.4$ \\
Voxel vote, $k=3$ & 0.535 & $+0.016$ [$+0.006$, $+0.027$] & $+1.3$ & $-5.9$ \\
Lesion vote, $k=2$ & 0.517 & $-0.002$ [$-0.015$, $+0.010$] & $-2.3$ & $+11.7$ \\
Lesion vote, $k=3$ & 0.531 & $+0.012$ [$+0.001$, $+0.024$] & $+1.1$ & $+4.6$ \\
\bottomrule
\end{tabular}
\end{table}

\subsection{Anatomic-region performance on the test cohorts}
\textbf{High-risk bone.} SAILS improved metastasis Dice within high-risk bone in both cohorts (Table~\ref{tab:organ}): from 0.552 to 0.576 on AutoPET~III ($+0.024$, 95\% CI $+0.006$ to $+0.047$) and from 0.653 to 0.667 on Deep-PSMA ($+0.014$, 95\% CI $+0.010$ to $+0.018$; both $p<0.01$). Metastasis Dice within all bone changed less ($+0.007$ and $+0.008$); both differences were significant by the rank-based test, but their 95\% CIs included zero, reflecting small gains in most cases and losses in a few. Inference-time fusion gave the largest bone and high-risk bone gains on Deep-PSMA ($+0.024$ and $+0.030$).

\textbf{Liver.} Liver metastases were too few for inference on AutoPET~III (2 evaluable cases). On Deep-PSMA (18 cases), metastasis Dice within the liver increased from 0.187 to 0.222 with SAILS, but the difference was not significant; fusion increased it to 0.231.

\begin{table}[ht]
\caption{Whole-body voxel-level performance on the test cohorts (disease-positive cases). Dice: mean $\pm$ SD. $\Delta$Dice: paired mean difference to the baseline with bootstrap 95\% CI. FNVol / FPVol: mean missed-lesion and false-positive volume per case (cc), with the paired difference to the baseline in parentheses. $p$: two-sided Wilcoxon signed-rank test vs.\ baseline (Holm-adjusted over Dice and FNVol). Inference-time fusion requires registering the atlas to every test patient; SAILS does not.}\label{tab:voxel}
\footnotesize
\setlength{\tabcolsep}{3pt}
\resizebox{\linewidth}{!}{%
\begin{tabular}{@{}llcccc@{}}
\toprule
Cohort & Method & Dice & $\Delta$Dice (95\% CI), $p$ & FNVol ($\Delta$, $p$) & FPVol ($\Delta$, $p$) \\
\midrule
AutoPET~III & Baseline & $0.520\pm0.292$ & --- & 12.14 & 5.47 \\
($n{=}110$) & Fusion & $0.526\pm0.292$ & $+0.007$ [$-0.002$, $+0.014$] & 10.99 ($-1.15$) & 7.14 ($+1.67$) \\
 & SAILS & $0.531\pm0.290$ & $+0.011$ [$+0.000$, $+0.022$], 0.006 & 9.46 ($-2.68$, $<$0.01) & 7.75 ($+2.28$, $<$0.001) \\
\midrule
Deep-PSMA & Baseline & $0.590\pm0.156$ & --- & 12.65 & 25.15 \\
($n{=}100$) & Fusion & $0.619\pm0.160$ & $+0.028$ [$+0.025$, $+0.032$] & 11.45 ($-1.21$) & 27.11 ($+1.96$) \\
 & SAILS & $0.582\pm0.153$ & $-0.009$ [$-0.021$, $+0.002$], 0.359 & 9.95 ($-2.70$, $<$0.01) & 43.36 ($+18.21$, $<$0.001) \\
\bottomrule
\end{tabular}
}
\end{table}

\begin{table}[ht]
\caption{Lesion-level detection on the test cohorts. Sens / PPV: lesion sensitivity and positive predictive value, pooled over all lesions in the cohort. +TP / +FP: total number of additional detected reference lesions and additional false-positive components in the cohort, relative to the baseline. $p$: two-sided Wilcoxon signed-rank test on per-patient sensitivity vs.\ baseline.}\label{tab:lesion}
\begin{tabular}{@{}llccc@{}}
\toprule
Cohort (reference lesions) & Method & Sens / PPV & $p$ (Sens) & +TP / +FP \\
\midrule
AutoPET~III & Baseline & 0.843 / 0.793 & --- & --- \\
(3,231) & Fusion & 0.859 / 0.762 & --- & 54 / 110 \\
 & SAILS & 0.857 / 0.759 & 0.004 & 46 / 150 \\
\midrule
Deep-PSMA & Baseline & 0.751 / 0.887 & --- & --- \\
(6,841) & Fusion & 0.770 / 0.876 & --- & 139 / 89 \\
 & SAILS & 0.769 / 0.865 & $<$0.001 & 126 / 198 \\
\bottomrule
\end{tabular}
\end{table}

\subsection{Computational Cost}
The atlas-weighted loss adds two element-wise operations per voxel to the cross-entropy and reads one precomputed atlas map per training case. Training time per epoch was 287~s for the baseline and 305~s for an atlas-weighted network on one NVIDIA L40S GPU; SAILS requires training three such networks. The atlas maps are warped to each training case once, offline. At inference, each SAILS member is a standard nnU-Net and takes 1.25--1.5~min per case, so SAILS takes 3.75--4.5~min per case when the three networks are run sequentially. No registration is performed. Inference-time fusion and the atlas-input model, in contrast, each require one deformable registration per test case ($\sim$20~min).

\subsection{Qualitative results}
\begin{figure}[ht!]
\centering
\includegraphics[width=\textwidth]{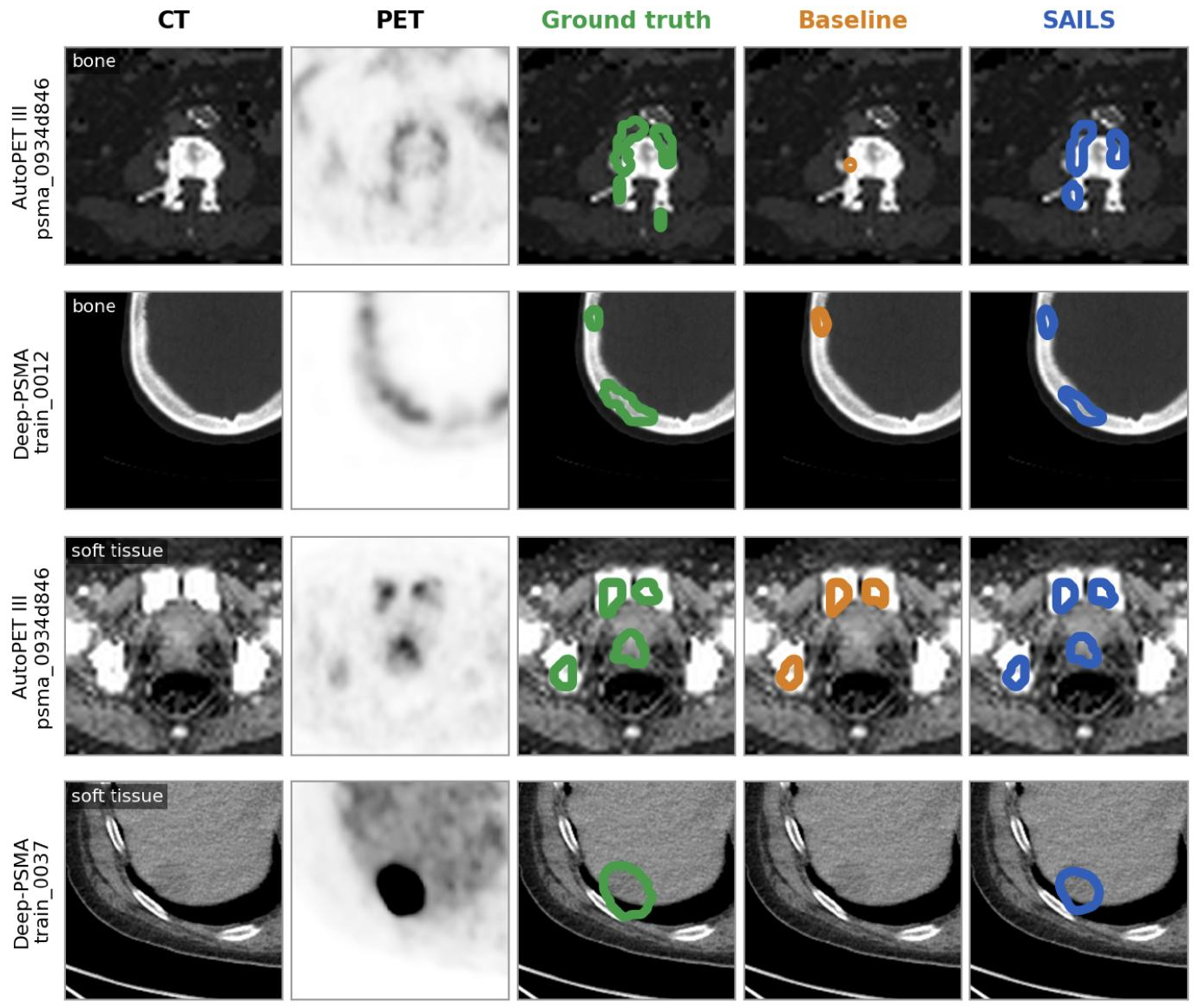}
\caption{Lesions recovered by SAILS. Each row shows one example lesion, at the center of the panel, that the baseline missed or segmented only minimally and that SAILS segmented more completely, for each anatomic region (tag in the CT panel) and test cohort. Columns: CT, PET (SUV 0--8, inverted gray), and the ground truth (green), baseline (orange), and SAILS (blue) outlined on the CT. Each row is an axial slice through the lesion. The number in the ground-truth panel is the volume of the example lesion. Other lesions visible in the panel are also outlined. Examples were selected based on improvement in lesion coverage and lesion volume, accounting for false-positive volume added within 4~cm, and illustrate the effect of the method rather than typical performance. Regions or cohorts without an eligible lesion have no row.}
\label{fig:qual_recovered}
\end{figure}

\begin{figure}[ht!]
\centering
\includegraphics[width=\textwidth]{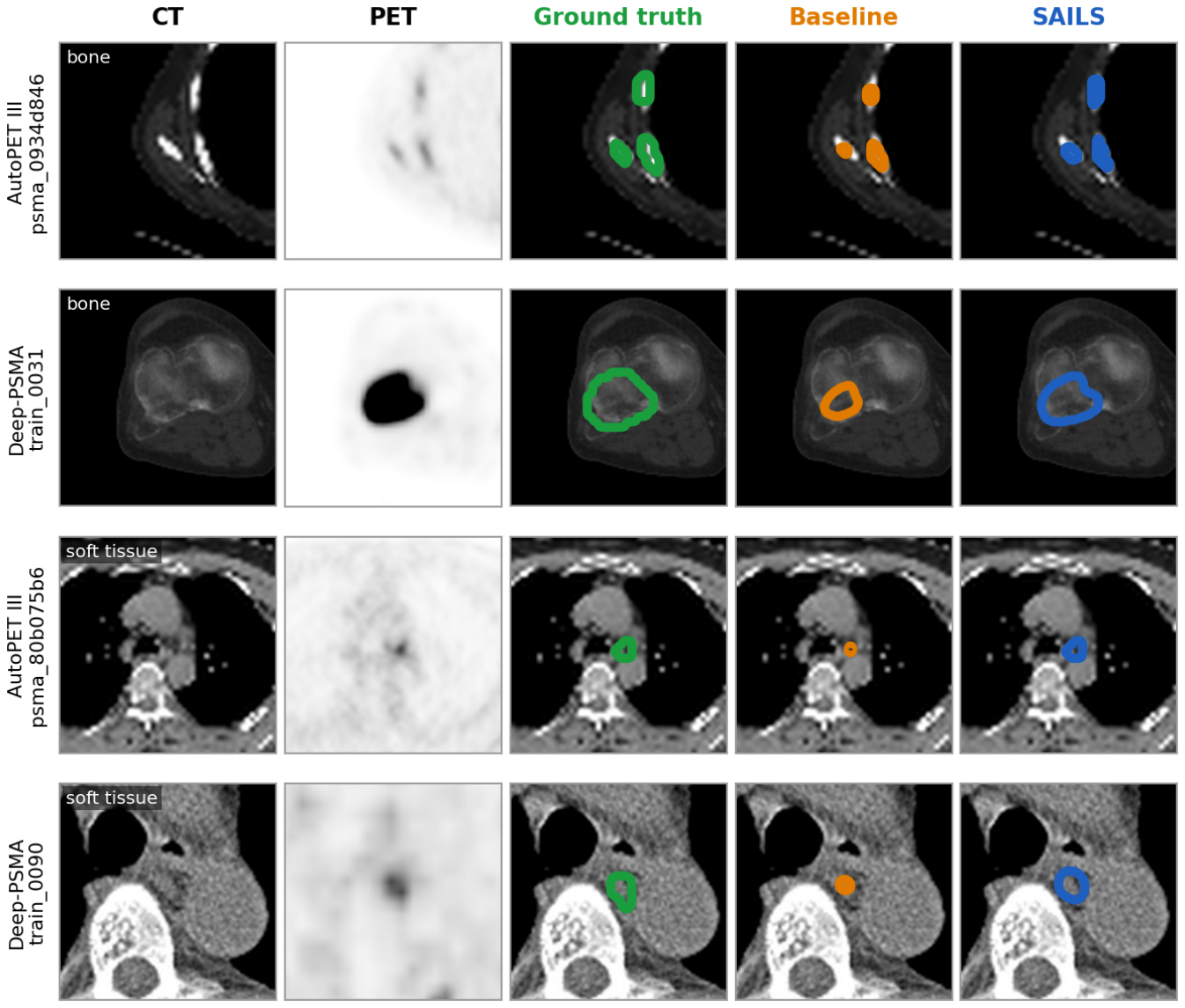}
\caption{Lesions refined by SAILS. Each row shows one example lesion that the baseline detected but under-segmented and that SAILS segmented more completely. Display conventions, lesion volumes, and example selection are as in Fig.~\ref{fig:qual_recovered}.}
\label{fig:qual_refined}
\end{figure}

Figs.~\ref{fig:qual_recovered} and~\ref{fig:qual_refined} show example bone and soft-tissue lesions from the AutoPET~III and Deep-PSMA test sets. In the recovered examples (Fig.~\ref{fig:qual_recovered}), SAILS segmented lesions that the baseline missed or segmented only minimally. These include bone lesions of 9.1 and 4.6~cc and soft-tissue lesions of 13.3 and 23.9~cc. In the refined examples (Fig.~\ref{fig:qual_refined}), the baseline detected the lesions but under-segmented them, whereas SAILS provided more complete segmentations, including a 1.8~cc mediastinal lesion on AutoPET~III.

\begin{table}[t]
\caption{Region-level metastasis Dice on the test cohorts: the Dice of the predicted and reference lesion masks restricted to voxels inside each region (not the Dice of the organ itself). Means are over cases evaluable for all three methods ($n$). Each cell gives the mean Dice and the paired difference to the baseline [95\% bootstrap CI]. $^{*}$Wilcoxon signed-rank $p<0.01$ vs.\ baseline. HR-bone: high-risk bone. Liver in AutoPET~III has too few cases (2) and is not tested.}
\label{tab:organ}

\centering
\small
\setlength{\tabcolsep}{3pt}
\renewcommand{\arraystretch}{1.2}

\begin{tabularx}{\linewidth}{@{}llcc
  >{\raggedright\arraybackslash}X
  >{\raggedright\arraybackslash}X@{}}
\toprule
Cohort & Region & $n$ & Baseline & Fusion & SAILS \\
\midrule
AutoPET~III & Bone & 86 & 0.553
  & 0.551 ($-0.001$ [$-0.020$, $+0.017$])
  & 0.560 ($+0.007$ [$-0.003$, $+0.017$])$^{*}$ \\
& HR-bone & 73 & 0.552
  & 0.574 ($+0.021$ [$+0.008$, $+0.035$])$^{*}$
  & 0.576 ($+0.024$ [$+0.006$, $+0.047$])$^{*}$ \\
& Liver & 2 & --- & --- & --- \\
\midrule
Deep-PSMA & Bone & 98 & 0.588
  & 0.613 ($+0.024$ [$+0.020$, $+0.028$])$^{*}$
  & 0.596 ($+0.008$ [$-0.004$, $+0.017$])$^{*}$ \\
& HR-bone & 91 & 0.653
  & 0.684 ($+0.030$ [$+0.026$, $+0.035$])$^{*}$
  & 0.667 ($+0.014$ [$+0.010$, $+0.018$])$^{*}$ \\
& Liver & 18 & 0.187
  & 0.231 ($+0.044$ [$+0.022$, $+0.069$])$^{*}$
  & 0.222 ($+0.035$ [$-0.021$, $+0.131$]) \\
\bottomrule
\end{tabularx}
\end{table}

\begin{figure}[ht]
\centering
\includegraphics[width=\textwidth]{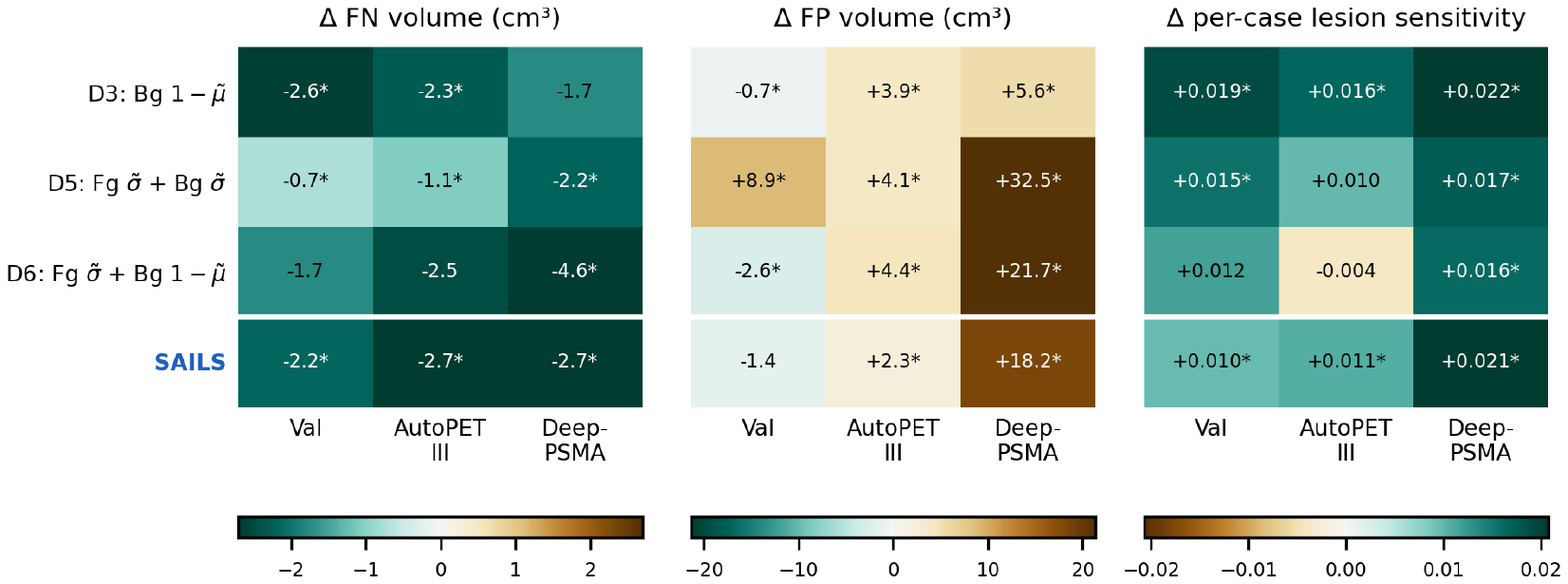}
\caption{Change in error type relative to the baseline for D3, D5, D6 and their ensemble (SAILS): false-negative volume, false-positive volume (cm$^3$ per case) and per-case lesion sensitivity. Teal marks improvement and brown marks worsening; each panel has its own symmetric color scale. $^{*}$Holm-adjusted $p<0.01$.}\label{fig:clinical}
\end{figure}

\section{Discussion}

We presented an atlas-weighted loss that brings a population-level metastasis atlas into the training of a whole-body PSMA PET/CT lesion segmentation network, and SAILS, a majority vote of three networks trained with different atlas weightings. The atlas enters only the loss, so every network is a standard nnU-Net and no atlas or registration is needed at inference. We evaluated SAILS on a held-out AutoPET~III test set and the external Deep-PSMA cohort at the voxel, lesion, and anatomic-region level.

The main effect of atlas-weighted training is on detection rather than whole-body overlap. Most atlas-weighted configurations reduced missed-lesion volume, while their effects on false-positive volume differed. The networks differed in which false positives they added, and majority voting allows us to take advantage of multiple predictions from different models. False positives produced by only one network are removed, whereas lesions found by at least two networks are kept. As a result, SAILS improved whole-body Dice in-distribution and detected additional reference lesions in both cohorts. On Deep-PSMA, which differs from the training data in tracer, scanner, annotation protocol, and disease burden, SAILS did not significantly change whole-body Dice. Metastasis segmentation within high-risk bone improved in both cohorts.

In practice, this behavior affects how the model can be used. When automated segmentation serves as a concurrent or second reader, a false-positive candidate is shown to the physician and can usually be dismissed as physiological uptake, whereas a missed lesion is not shown at all and is easily overlooked by a reader at the same site. Errors of the two kinds are therefore dissimilar in terms of clinical cost. The sites where detection improved most consistently, high-risk bone lesions at weight-bearing joints, the junctional spine and the posterior elements, are those for which prophylactic radiotherapy reduces skeletal-related events \citep{Gillespie24HighRiskBone}. In oligometastatic disease, lesion sensitivity is critical for deciding eligibility for metastasis-directed therapy \citep{alberto2022role,phillips2020outcomes}. On Deep-PSMA, a pre-radioligand-therapy cohort with high disease burden, SAILS detected 126 additional reference lesions. The remaining additional false positives were mostly small (median 0.13--0.54~cc), which limits their effect on total tumor volume estimates that feed survival models and nomograms \citep{Seifert20TTV,Seifert23RiskScore,Gafita21Nomograms}. They still add reading time, however, and reducing them further is necessary before clinical use.

Unlike segmentation studies that rely on voxel-level metrics alone, we report lesion-level and anatomic-region-level results relevant to treatment planning, including high-risk bone and liver, where missed lesions have the largest clinical consequence. The region analysis shows that improvements in whole-body Dice and in clinically weighted regions can diverge. On Deep-PSMA, whole-body Dice did not improve while metastasis Dice within high-risk bone rose.

SAILS and inference-time fusion act at different points and can be complementary rather than competing. In particular, fusion adjusts predictions that the network has already made and requires registration at test time. On Deep-PSMA, fusion improved Dice and reached similar lesion sensitivity with far fewer additional false positives (Table~\ref{tab:lesion}), at the cost of one deformable registration per patient. SAILS trades some specificity for independence from registration. Applying fusion to the SAILS output, which would combine training-time and inference-time use of the atlas, is a natural next step.

In an exploratory run at 500 epochs, the difference between the Fg~$\tilde\sigma$ ($\alpha=1$) model (D2) and the baseline was larger than at 300 epochs on AutoPET~III ($+0.046$ versus $+0.009$ Dice), with less false-positive volume rather than more ($-8.0$ cc). It was not larger on the external Deep-PSMA cohort ($-0.025$). This suggests that the up-weighted objective may need a longer schedule than the baseline to converge, consistent with the more complex objective function and gradient when compared to vanilla cross-entropy. Thus, a comparison with the same models at longer schedules is warranted. Note that this run used D2 rather than the SAILS ensemble predictions and is reported as exploratory only.

The atlas-weighted loss is designed for practical use, as it depends only on the network's softmax output, the reference labels, and the atlas. Thus, in principle, it can replace the loss of other segmentation pipelines without architectural changes, and the atlas is built once, offline. The voting step only needs the members' binary predictions.

\subsection*{Limitations}\label{sec:lim}
Our study has several limitations. First, we evaluated a single network architecture and one registration pipeline, so the benefit may differ for other architectures or registration methods. Second, each configuration was trained once with a single seed, so run-to-run variability was not measured, and part of the benefit of voting may come from ensembling itself; an ensemble of baseline networks trained with different seeds was not evaluated. Third, all models were trained for 300 epochs, which may under-train the atlas-weighted objective relative to the baseline. Fourth, SAILS still increased false-positive volume, particularly on the external cohort, and requires three forward passes at inference. Finally, the atlas was built from a single data source and encodes only a marginal spatial prior, without conditioning on patient-level factors such as disease stage; richer priors remain future work. For training cases the atlas average includes the case's own lesion mask, which makes the weight more informative about that case at training time than at test time.

\section{Conclusion}\label{sec:conclusion}
We introduced an atlas-weighted loss that uses a population atlas of prostate cancer metastases to guide whole-body PSMA PET/CT segmentation, and SAILS, a majority vote of three networks trained with it. The atlas reweights the cross-entropy only where the network is uncertain, so each network keeps the standard nnU-Net architecture, and no atlas or registration is required at inference. On the held-out AutoPET~III test set, SAILS improved whole-body Dice, missed less lesion volume, and detected more lesions (pooled sensitivity 0.843 to 0.857). On the external Deep-PSMA cohort, it detected 126 additional lesions without a significant change in whole-body Dice. Metastasis segmentation within high-risk bone improved in both cohorts. These lesions, if missed, matter for treatment planning. The main limitation is a remaining increase in false-positive volume, mostly due to small components. Atlas-guided training with majority voting is therefore a simple, registration-free way to make a segmentation network more sensitive to clinically important lesions, and combining it with inference-time fusion is a natural next step.

\newpage
\bibliographystyle{unsrtnat}
\bibliography{cas-refs}

\end{document}